\documentclass[5pt,two column]{article}
\usepackage{graphicx}   
\usepackage{multirow}
\usepackage{subfigure}
\usepackage{geometry}
\usepackage{stfloats}
\usepackage{booktabs}
\usepackage{amsfonts}
\usepackage{amsmath}

\usepackage{titlesec}
\titleformat{\section}%设置section的样式
{\raggedright\normalsize\bfseries}%右对齐，4号字，加粗
{\thesection .\quad}%标号后面有个点
{0pt}%sep label和title之间的水平距离
{}%标题前没有内容

\titleformat{\subsection}%设置section的样式
{\raggedright\small\bfseries}%右对齐，4号字，加粗
{\thesection .\quad}%标号后面有个点
{0pt}%sep label和title之间的水平距离
{}%标题前没有内容

\usepackage{caption}
\usepackage{mathrsfs}
\begin{document}\small

\captionsetup{font={small}}
\title{Model Embedded DRL for Intelligent Greenhouse Control}
\author{Tinghao Zhang, Jingxu Li, Jingfeng Li, Ling Wang, Feng Li, Jie Liu\\
IoT and Ambient Intelligence Lab\\
Harbin Institute of Technology, China
}
\date{}

\maketitle
\begin{abstract}
\begin{quote}
Greenhouse environment is the key to influence crops production. However, it is difficult for classical control methods to give precise environment setpoints, such as temperature, humidity, light intensity and carbon dioxide concentration for greenhouse because it is uncertain nonlinear system. Therefore, an intelligent close loop control framework based on model embedded deep reinforcement learning (MEDRL) is designed for greenhouse environment control. Specifically, computer vision algorithms are used to recognize growing periods and sex of crops, followed by the crop growth models, which can be trained with different growing periods and sex. These model outputs combined with the cost factor provide the setpoints for greenhouse and feedback to the control system in real-time. The whole MEDRL system has capability to conduct optimization control precisely and conveniently, and costs will be greatly reduced compared with traditional greenhouse control approaches.
\end{quote}
\end{abstract}

\section{Introduction}
Greenhouses have been widely used in agriculture, especially in norther part of China, because of its high efficiency and less weather dependency. Traditional greenhouse management requires operators to monitor crops status by going to the greenhouses in person. Besides, operators can hardly offer optimization control for the greenhouse environment (GE) until they have accumulated enough experience about the crops growth. Once the environment of greenhouse changes, previous experience may not be effective any longer. Therefore, traditional control methods are neither manpower nor financial efficiency. How to regulate the greenhouse conveniently and efficiently has attracted many researchers’ attention. Recent years, fast development of internet of things (IoT) allows to supervise the greenhouses by remote monitor\cite{Li14,Kodali16,Reka19}. Data collected by sensing network are uploaded to cloud platform, where it can be monitored in real time. At the same time, operators can remotely control the GE.

In terms of control methods, expert systems can make decisions like human and were used to control the environment for certain crops\cite{Kano88,Tchamitchian06}. These systems only considered one or two environmental factors, hence limiting its accuracy of GE control. Recently some researchers have applied popular control algorithms like proportional integral differential (PID) algorithm into greenhouse control\cite{Wang15,Chaudhary19}, while others have trained ANN with the help of other computing algorithms and gotten the control models for GE\cite{Luan11,Alejandro17}. Literature in \cite{Taki18} used machine learning to estimate the temperatures, energy lost and exchange based on outdoor environment. However, few literatures have given suggestions on setpoints determination for greenhouse. What’s more, none of them has considered the relation between crop growth and GE control. Specifically, the demand of crop growth on the environment can be various under different conditions. What’s more, the selection of setpoints for GE is a trade-off between crop growth and costs to achieve optimization choice for smart greenhouse cultivation systems.

In this paper, we propose to apply MEDRL to determine the setpoints for GE. Embedded models are trained to forecast the crop growth so that DRL can analyze complex cultivation systems more deeply and balance crop growth and costs when making decisions. The ideas of growth modeling are greatly adaptive and can be used in various kinds of crops. Besides, image processing algorithms are employed to extract specific features of crops, which will be a part of input for DRL models, from images of crops. In order to formulate MEDRL models as most accuracy as possible, pattern recognition based on computer vision techniques is adopted to identify growing periods and other differences (e.g. sex) of the crops. Crops will have different MEDRL models based on these differences. For implementation, we designed a hardware framework with three feasible schemes for choosing control devices. In conclusion, the MEDRL-based greenhouse environment control system combined with crop growth embedded modeling is proposed in this paper, and also being implemented in the reality.

\begin{figure*}[h]
  \centering
  \includegraphics[width=16.0cm]{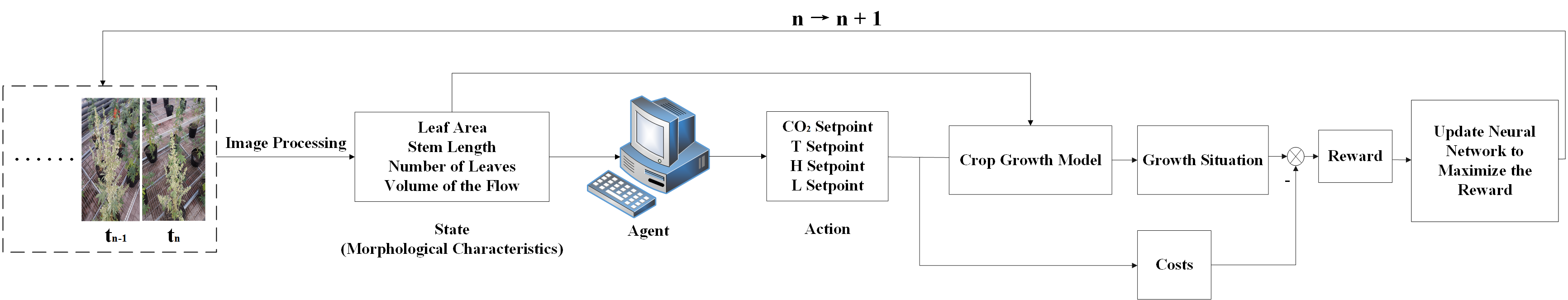}
\caption{MEDRL Framework for Greenhouse Environment Control}
\label{DRL}
\end{figure*}

\section{Framework of MEDRL Model}
\subsection{DRL model for Greenhouse Environment}
The environment in greenhouse is very complex, so even experts in this field don’t have enough confidence to decide the most reasonable setpoints for GE. Advanced techniques are necessary for deeply understanding greenhouse systems. 

Reinforcement learning (RL), a novel machine learning algorithm, has been proved to be effective in optimization control\cite{Sutton92}. DRL combines the merits between deep learning (DL) and RL\cite{Mnih15}, and has been successfully applied into many research areas such as autonomous driving\cite{Hilleli18}, game playing\cite{Devendra17}, decision making problems\cite{Lyu19} and so on. Therefore, DRL is applied in this paper to analyze the greenhouse system and provide setpoints for each environment factors.

\iffalse
To be more specific, RL can solve sequential decision problems that usually are formulated as Markov decision processes (MDPs). For instance, the RL agent obtains a state $s_t$ at a time step $t$ and then generates an action $a_t$. Reward $r_t$ and next state $s_{t+1}$ are obtained at the same time. Meanwhile, cumulative discounted reward $R_t$ and the action-value function $Q^\pi(s_t,a_t)$ with policy $\pi$ are defined as following: 
\begin{equation}
  R_t=\Sigma^T_{i=t}{\gamma^{i-t}r(s_i,a_i)}
\end{equation}
\begin{equation}
  Q^\pi(s_t,a_t) = \mathbb{E}_\pi\left[ R_t|s_t,a_t \right]
\end{equation}
where $E$ is the expectation of the probabilities. The agent continually takes actions until an episode is over. The final goal of RL algorithms is to obtain a policy $\pi$ that maximizes the expected cumulative discounted rewards from the initial position $R = \mathbb{E}\left[ \Sigma^T_{t=0}{\gamma^{t}r(s_t,a_t)} \right]$.
When the policy for collecting training data is the same policy network that is being learned, such RL algorithms are called on-policy. Otherwise are called off-policy. Compared with on-policy RL, Off-policy RL algorithms gain more attention from researchers because of their higher training efficiency and better exploration. In addition, the action space in RL can be either discrete or continuous. RL algorithms with discrete action space are easy to train while having lower precision, as opposed to the models with continuous action space.
\fi

In order to apply DRL in smart greenhouse system, three aspects of information are needed to be defined: state, action and reward function (The reward function). As mentioned above, the state is the input of the DRL model and should describe the controlled environment as accurately as possible. Action represents what the agent will do after receiving the state. The reward function is to evaluate this action.

In this paper, the state is the morphological characteristics of the crops, such as leaf area, stem length, number of leaves and volume of flow. Extracting these information from the crop images may require image processing algorithms. In the respect of the action, four physical variables, including carbon dioxide levels, light intensity, temperature and humidity, are taken into consideration. The reward function contains two parts: growth situation and corresponding costs,
\begin{equation}
  RF=a*GS-b*C
\end{equation}
where $RF$ denotes the reward function, $C$ is the costs, $GS$ is growth situation, $a$ and $b$ are the coefficients. Intuitively, it is a trade-off problem to decide the values of $a$ and $b$.

It should be noted that crops have different growing periods. For example, Industrial hemp, has four growing periods during its life, germination, seedling, mature, and blooming stages. For those periods that haven’t bloom yet, the sum of stem length and the number of leaves is regarded as the growth situation. In contrast, the volume of the flower represents the growth situation. Costs mainly come from electricity bills when controlling sensors, cameras and heaters. Generating carbon dioxide or controlling air flow also has costs. The values of $a$ and $b$ can be determined by further tests. Figure~\ref{DRL} shows the framework of the DRL control for GE, where T means temperature, H means humility, and L means light intensity.
\begin{figure}[htb]
  \centering
  \includegraphics[width=8.0cm]{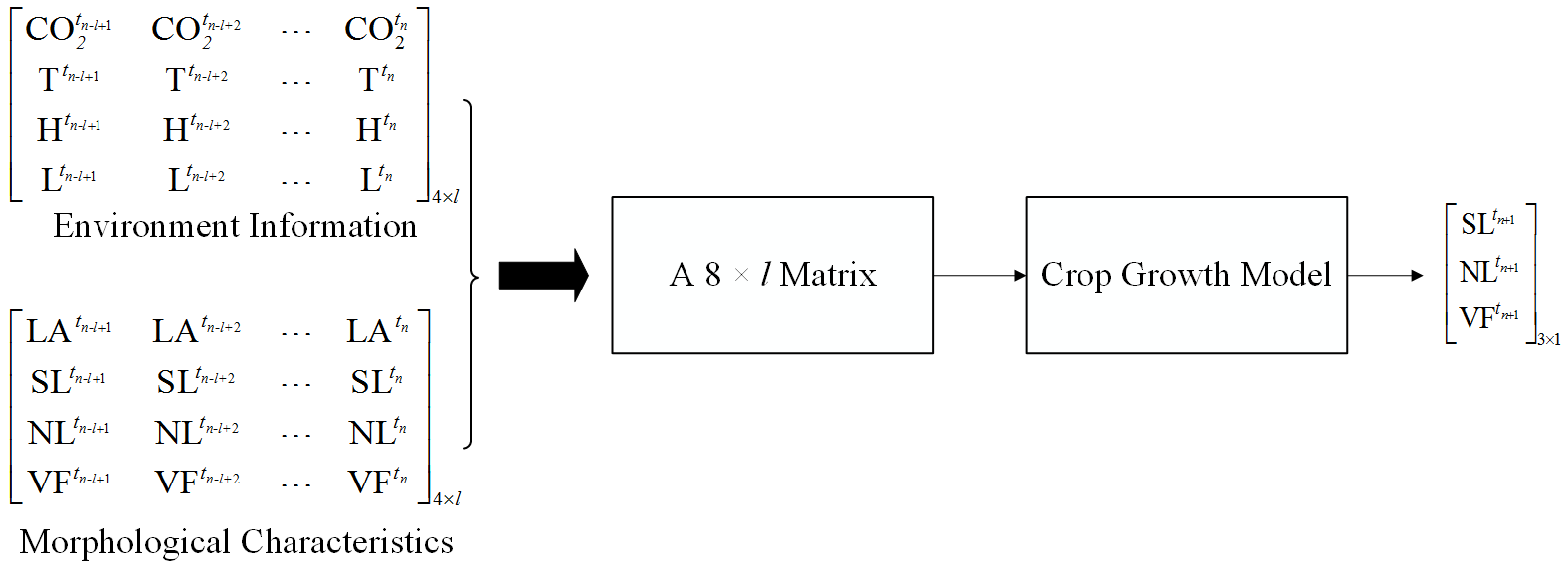}
\caption{Training the Crop Growth Model}
\label{Simu}
\end{figure}

\subsection{Crop Growth Modeling}
As is shown in Figure~\ref{DRL}, a crop growth model is embedded in the DRL model. Specifically, the agent gives the setpoints for each environment variable during the DRL training period, and the growth situation can be obtained by a embedded crop growth model. Then the growth situation combined with costs is used to calculate the reward value. However, most simulation approaches for the crop growth are based on expert models, which need rich knowledge about agriculture and are not universal for different kinds of crops. Therefore, we resorted to ML algorithms to forecast the crop growth under different crop environments.

\begin{figure}[h]
\centering
\subfigure[]{
\begin{minipage}[t]{0.5\linewidth}
\centering
\includegraphics[width=4cm]{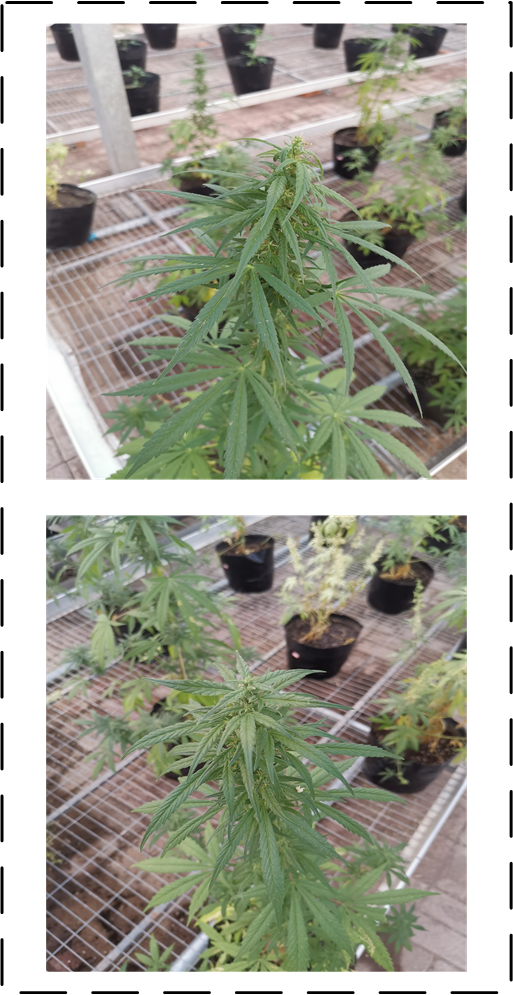}
%\caption{fig1}
\end{minipage}%
}%
\subfigure[]{
\begin{minipage}[t]{0.5\linewidth}
\centering
\includegraphics[width=4cm]{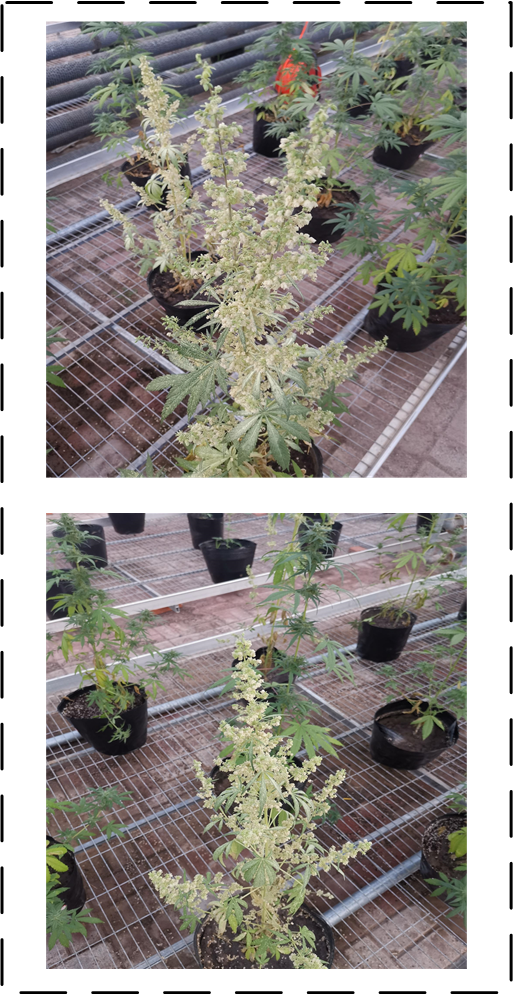}
%\caption{fig2}
\end{minipage}%
}%
\caption{Industrial Hemp in Different Growing Periods: (a) Female Plants; (b) Male Plants.}
\label{plant}
\end{figure}

\begin{figure*}[b]
  \centering
  \includegraphics[width=16.0cm]{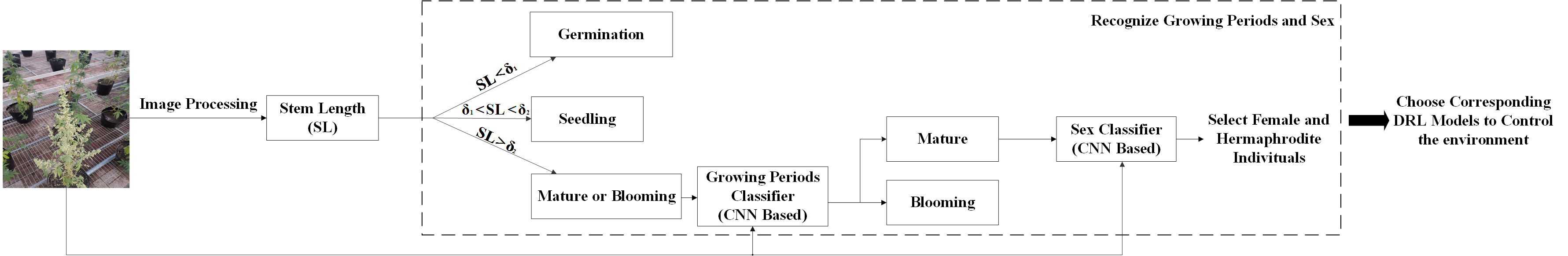}
\caption*{Figure 4: Growing Period and Sex Classification using CNN}
\label{AI}
\end{figure*}

Specifically, the collected data are divided into different classes based on their growing periods, and each growing period has its own growth model. Because the growth situation is required after the agent takes an action, the input of the model is supposed to have the same variables as the action has. Morphological characteristics are also be included as input. Moreover, the crop growth usually takes a long time, and not only current environment but also the historical environment may have influence on it. Therefore, the input can have both current and historical information of the crops. Besides traditional ML algorithms, recurrent neural network (RNN) and its improvement like long short-term memory network (LSTM) are also a good choice for us to train the model, as it can analyze the data on the current and past time.

Figure~\ref{Simu} shows the way to train the model, where $l$ denotes the number of samples that used to train the model in one step, $n = l,l+1,...,N$ and $N$ denotes the total number of the samples. The loss function is defined as the mean square error between the true and predicted values. With the help of embedded models, DRL can be trained in a more reasonable way.

\section{Growing Period and Sex Classification}

Unique DRL and growth models should be formulated for different growing periods, as crops physiology varies. In addition, sex classification also plays an important role in practice. Take industrial hemp as an example, female and hermaphrodite crops are known to have much higher economic value than male ones so it’s crucial to control the ratio of male and female crops. Therefore, growing periods and sex classification are very necessary to provide precise control and increase the production.

However, it’s difficult for people without professional experience to recognize growing periods. For instance, Figure~\ref{plant} is the photos of the industrial hemp under plastic shed. Most people would feel that the crops with the same sex in different GPs have little difference. Therefore, we apply deep learning (DL) techniques to conduct this pattern recognition task automatically. 

Convolutional Neural Network (CNN) has been widely used and has shown a better performance over many traditional ML algorithms in the computer vision area, so we apply CNN to conduct growing periods and sex classification. Figure 4 is a framework of the recognition process, where $\delta_1$ and $\delta_2$ are two thresholds. Sex of industrial hemp can be known in the mature stage, so classification is conducted during that period. Noted that for some kinds of crops like industrial hemp, the seedling period and germination period can be basically recognized by stem length. Therefore, we simplify the task into a binary classification problem, which saves the training costs without reducing the recognition rate.

\section{Hardware Configuration}

We designed a overall hardware framework for implementing proposed control system as shown in Figure 5. Environment factors including carbon dioxide, light intensity, temperature and humidity are collected by sensors and sent to the microcontroller unit (MCU). Raspberry Pi (RPi) is the connection between MCU and the platform. After receiving the data, MCU sends the data to the cloud platform through RPi. Images of the crops collected by cameras will be sent to RPi directly. Image processing, DRL control, growing periods classification are all conducted in the cloud platform. The setpoints given by DRL models are in turn received by MCU through RPi. Based on the current intensity and types of devices, we came up with three types of control devices. Four types of controllers will then tune the GE to reach the setpoints. In addition, sensors network configuration is shown in Figure 6. The connection approaches between different sensors and devices include I2C, serial port protocol, and Wifi or IoT. 

\begin{figure*}[htb]
  \centering
  \includegraphics[width=16.0cm]{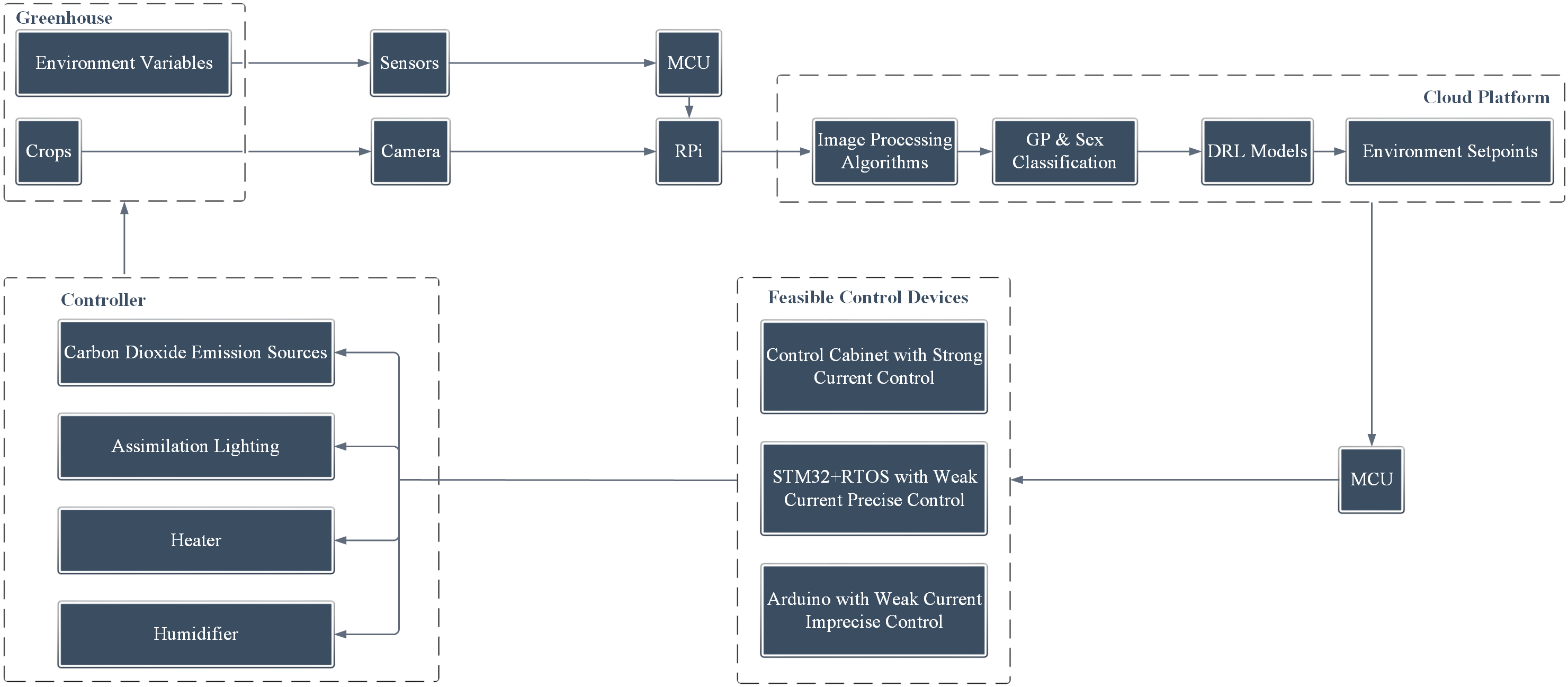}
\caption*{Figure 5: Overall Hardware Framework}
\label{hard}
\end{figure*}

\begin{figure}[htb]
  \centering
  \includegraphics[width=8.0cm]{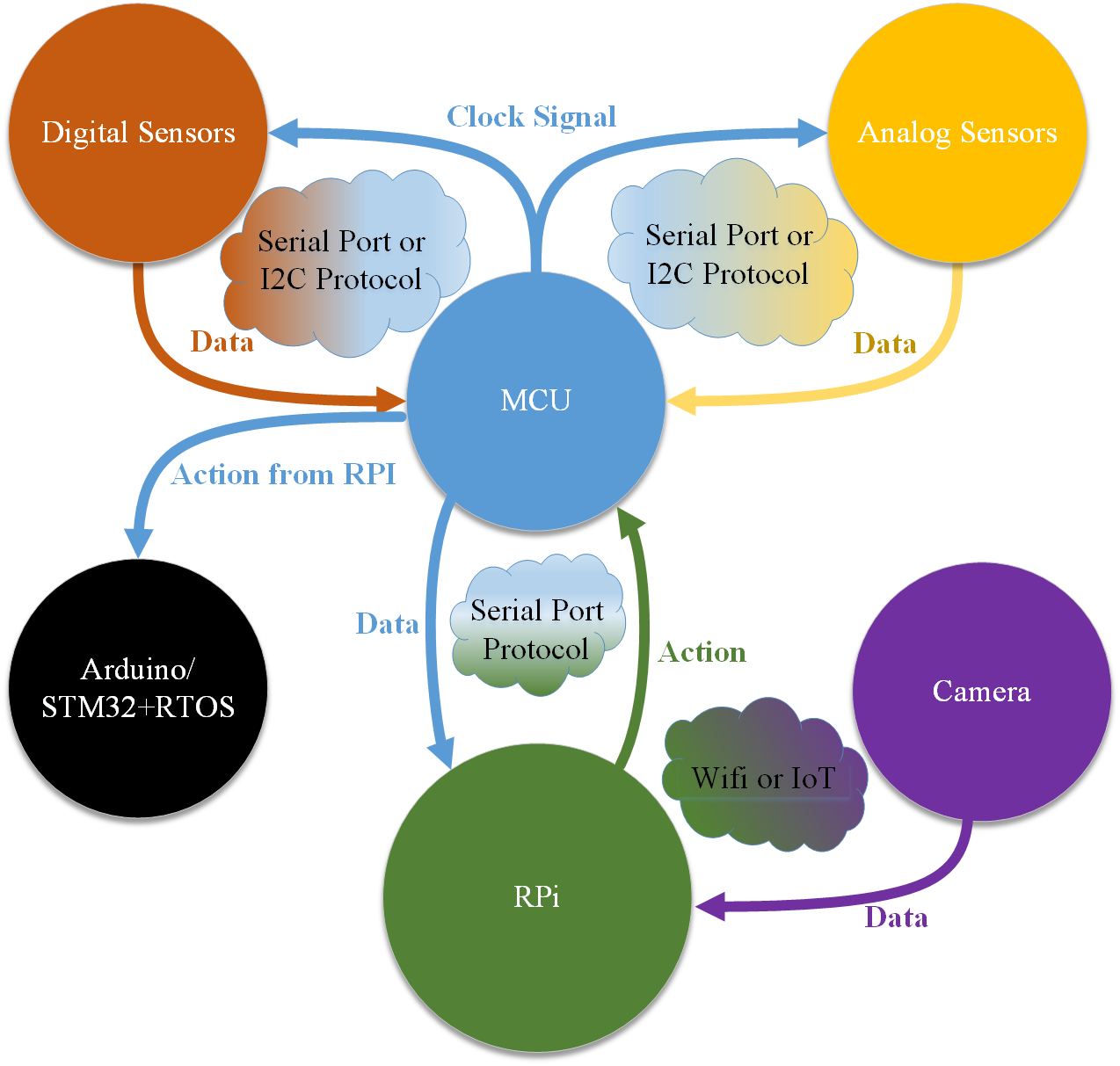}
\caption*{Figure 6: Sensor Network Architecture}
\label{Sensor}
\end{figure}

Precise control is required after DRL offers the setpoints for GE. A mixture of three schemes are used in this control system.

\begin{itemize}
    \item \textbf{Control Cabinet with Strong Current Control}: Strong current control has high stability and technological maturity. Large electric fans, heaters and artificial illuminators can also be driven by it. However, it is not easy to realize intelligent control by related algorithms. 
    \item \textbf{STM32 \& RTOS with Weak Current Precise Control}: STM32 with RTOS is very precise, but the capability to drive high power loads is very poor. In this case, most controllers are low powered and control circuit will be complicated. 
    \item \textbf{Arduino with Weak Current Imprecise Control}: The precision of Arduino is not as high as that of STM32. But it’s easy for people to learn it because of its open source.
\end{itemize}

The hardware framework is currently being built based on above schemes.

\section{Conclusion}
In this work, a model embedded deep reinforcement learning (MEDRL) control framework was proposed for intelligent greenhouse environment. The embedded crop growth models were formulated using ML algorithms, which enjoy more generalization ability than traditional experts models and can be simply migrated in many kinds of crops. Besides, computer vision algorithms were applied to recognize, and we will formulate crop growth models and DRL models for the crops with different growing periods and sex, respectively. It was a refinement of the modeling process and also enhances reliability for DRL control. A hardware framework is being implemented in a greenhouse environment with industrial hemp growing. The whole system has a great potential to control the costs of greenhouse production, hence a good application in real life.

\footnotesize
\bibliographystyle{acm}
\bibliography{samplebase}

\end{document}